\documentclass[runningheads,a4paper]{llncs}
\usepackage{wrapfig}
\usepackage{graphicx}
\usepackage{booktabs}
\usepackage{colortbl}
\usepackage{subcaption}
\usepackage{hyperref}
\hypersetup{
hidelinks
}
\usepackage{amsmath}
\usepackage{multirow}
\usepackage{enumitem}
\usepackage{cite}
\usepackage{amssymb}
\usepackage{graphicx}
\usepackage{amsmath} 
\usepackage{url}
\usepackage{algorithm}
\usepackage{algorithmic}
\usepackage{marvosym}

\newcommand{\keywords}[1]{\par\addvspace\baselineskip
\noindent\keywordname\enspace\ignorespaces#1}

\begin{document}

\mainmatter  

\title{PhysMamba: Efficient Remote Physiological
Measurement with SlowFast Temporal Difference Mamba}

\titlerunning{PhysMamba}

%
%
\author{Chaoqi Luo\thanks{Equal contribution. \quad \textsuperscript{(\Letter)} Zitong Yu is the corresponding author. } \and Yiping Xie\footnotemark[1] \and Zitong Yu\textsuperscript{(\Letter)}}

%
\authorrunning{PhysMamba}

\institute{School of Computing and Information Technology, Great Bay University}

%
%

\toctitle{Lecture Notes in Computer Science}
\tocauthor{Authors' Instructions}
\maketitle

\begin{abstract} 
Facial-video based Remote photoplethysmography (rPPG) aims at measuring physiological signals and monitoring heart activity without any contact, showing significant potential in various applications. Previous deep learning based rPPG measurement are primarily based on CNNs and Transformers. However, the limited receptive fields of CNNs restrict their ability to capture long-range spatio-temporal dependencies, while Transformers also struggle with modeling long video sequences with high complexity. Recently, the state space models (SSMs) represented by Mamba are known for their impressive performance on capturing long-range dependencies from long sequences. In this paper, we propose the PhysMamba, a Mamba-based framework, to efficiently represent long-range physiological dependencies from facial videos. Specifically, we introduce the Temporal Difference Mamba block to first enhance local dynamic differences and further model the long-range spatio-temporal context. Moreover, a dual-stream SlowFast architecture is utilized to fuse the multi-scale temporal features. Extensive experiments are conducted on three benchmark datasets to demonstrate the superiority and efficiency of PhysMamba. The codes are available
at \href{https://github.com/Chaoqi31/PhysMamba}{\textcolor{blue}{Link}}.

\end{abstract}

\vspace{-2.8em}
\section{Introduction}
\vspace{-0.5em}
Remote photoplethysmography (rPPG) is a non-invasive technology designed to measure physiological signals such as heart rate (HR) and heart rate variability (HRV) by capturing subtle changes in blood volume from a distance. Unlike traditional methods like electrocardiography (ECG) and photoplethysmography (PPG), which require direct skin contact, rPPG uses standard cameras to detect variations in light absorption and reflection due to blood flow, providing a more convenient and comfortable monitoring solution.

In the early stages of rPPG development, facial video analysis became a focal point for extracting physiological signals. Researchers employed traditional signal processing techniques to track color changes in specific regions of interest (ROIs) on the face\cite{POS,CHROM,ICA,LCI,green}. These methods aimed to isolate the periodic signals corresponding to the cardiac cycle. Despite their innovative approach, these techniques often struggled with accuracy due to noise from environmental light fluctuations, facial movements, and other external factors. 


In recent years, the application of deep learning networks has revolutionized the field of facial rPPG measurement. CNNs and transformer-based architectures have been employed to enhance the rPPG signals reconstruction from facial videos\cite{Deepphys1, PhysNet, Efficientphys1, Physformer++, Rhythmformer, TransRPPG, Spiking-physformer}. However, CNNs are efficient in extracting local spatial features, while struggling with capturing long-range dependencies and temporal context. On the other hand, although the Transformer's self-attention mechanism achieves global context capture, it encounters difficulties in focusing on relevant local information when handling long video sequences.


Recently, the state space models (SSMs)\cite{SSM1, SSM2, SSM3}, especially Mamba\cite{SSM1} with its selective scan mechanism that allow models to dynamically select relevant information based on the input, which preserves earlier information while integrating recent information, has emerged as an efficient model to capture long-range dependencies when dealing with long sequences. The excellent long-range modeling capacity of SSMs motivates us to exploring the potential of Mamba for facial rPPG measurement task. In this paper, we propose a Mamba-based model PhysMamba. Specifically, we introduce a Temporal Difference Mamba (TD-Mamba) block which integrates temporal forward and backward Mamba (Bi-Mamba) with Temporal Difference Convolution (TDC) for efficiently capturing long-range spatio-temporal dependencies based on the refined fine-grained local temporal dynamics. Moreover, channel attention (CA) is also included in the block to reduce channel redundancy. Simultaneously, we utilize a dual-stream SlowFast architecture to fuse crucial multi-scale physiological features. 

The main contributions can be summarized as follows:
\begin{itemize}[label=$\bullet$]
  \item We propose the PhysMamba, a Mamba-based framework to leverage the Temporal Difference Mamba (TD-Mamba) block to enhance long-range spatio-temporal dependencies capture based on fine-grained temporal difference clues aggregation.
  \item A dual-stream SlowFast architecture is utilized for effective integration of multi-scale temporal features to reduce temporal redundancy while maintaining fine-grained temporal clues.
  \item Extensive experiments conducted on three benchmark dataset demonstrate that the proposed PhysMamba achieves superior performance and efficiency compared to previous CNN- and Transformer-based approaches.
\end{itemize}

\vspace{-1.8em}
\section{Related Work}
\vspace{-0.8em}
\noindent\textbf{Remote Photoplethysmography Measurement.} \quad
Traditional approaches for rPPG measurement have predominantly relied on analyzing periodic signals in facial regions of interest (ROI) by signal processing methods\cite{POS,CHROM,ICA,LCI,green}. In recent years, the advent of deep learning methods were introduced to rPPG measurement task. Convolutional neural networks (CNNs) have been employed for both skin segmentation and rPPG feature extraction. Some early approaches utilized 3D CNNs or 2D CNNs to capture spatial-temporal information for rPPG signals reconstructing\cite{PhysNet, Deepphys1,TS-CAN, Efficientphys1, rPPGNet}. More recently, transformers are utilized to enhance quasi-periodic rPPG features and global spatio-temporal perception\cite{Physformer, Physformer++, Rhythmformer, TransRPPG, Spiking-physformer}. However, Mamba-based rPPG measurement is rarely explored.

\vspace{0.3em}
\noindent\textbf{State Space Models.} \quad
Recently, State-Space models (SSMs)\cite{SSM1, SSM2}, particularly structured state space sequence models (S4)\cite{SSM1}, have emerged as an effective class of architectures for long sequence modeling. These models can be considered as an integration of recurrent neural networks (RNNs) and convolutional neural networks (CNNs). Mamba\cite{Mamba} further introduced a selective mechanism using parallel scan based on S4, allowing the model to select relevant information in an input-dependent manner. A series of studies have shown superior performance with SSM-based models on vision tasks such as classification\cite{VisionMamba}, video understanding\cite{VideoMamba}, and segmentation\cite{U-Mamba}. Inspired by this, we explore the capacities of Mamba for long-range spatio-temporal modeling on rPPG measurement.


\vspace{-1.0em}
\section{Methodology}
\vspace{-0.5em}
\subsection{Preliminaries}
\vspace{-0.3em}
\subsubsection{State Space Models (SSMs)} are foundational systems in control theory, used to model dynamic systems through state-space representation. SSMs can map a 1D-dimensional function or sequences $ x(t) \in \mathbb{R}^L \rightarrow y(t) \in \mathbb{R}^N $ through a hidden state $h(t) \in \mathbb{R}^N $, which typically described by the a following continuous linear time-invatiant (LTI) system of Ordinary Differential Equations (ODEs):
\begin{equation} \small
\begin{aligned}
h'(t) &= A h(t) + B x(t), \\
y(t) &= C h(t), 
\end{aligned}
\end{equation} 
where $A \in \mathbb{R}^{N \times N}$, $B \in \mathbb{R}^{N \times 1}$ and $C \in \mathbb{R}^{1 \times N}$ are learnable parameters. To integrate this continuous systerms into deep learning algorithms, Mamba\cite{SSM1} uses discretization methods. Specifically, a time-scale $\Delta $ is employed to convert continuous parameters $A$ and $B$ into discrete parameters $\bar A $ and $\bar B$ using the Zero-order hold (ZOH) method, defined as follows:
\begin{equation} \small
\begin{aligned}
\bar{A} &= \exp(\Delta A), \\
\bar{B} &= (\Delta A)^{-1} (\exp(\Delta A) - I) \cdot \Delta B. 
\end{aligned}
\end{equation}

The discretized form of the continuous ODEs transforms the model into a linear recurrent mode for efficient inference where the inputs are considered one timestep at a time. This is expressed as:
\begin{equation} \small
\begin{aligned}
h(t) &= \bar A h(t-1) + \bar B x(t), \\
y(t) &= C h(t). 
\end{aligned}
\end{equation}

Moreover, the model can be also computed in a global convolution way for efficient parallelizable training, which can be represented by:
\begin{equation} \small
\begin{aligned}
\bar{K} &= (C \bar{B}, C \bar{A} \bar{B}, \ldots, C \bar{A}^{L-1} \bar{B}, \ldots), \\
y &= x \ast \bar{K},
\end{aligned}
\end{equation}
where $L$ denotes the length of the sequence $x$, $\bar K \in \mathbb{R}^{L}$ denotes the convolution kernel and $*$ represents the convolution operation.

Compared with the traditional time- and input-invariant SSMs, the recent powerful state space model, Mamba, utilizes an input-dependent Selective Scan Mechanism (S6) to allow the parameters $\Delta \in \mathbb{R}^{B\times L \times D} $, $\bar B \in \mathbb{R}^{B\times L \times N} $ and $C \in \mathbb{R}^{B\times L \times N} $ are derived from the input data $x \in \mathbb{R}^{B\times L \times D} $.

\begin{figure}[t]
\vspace{-1.5em}
\centering
\includegraphics[width=1.0\textwidth]{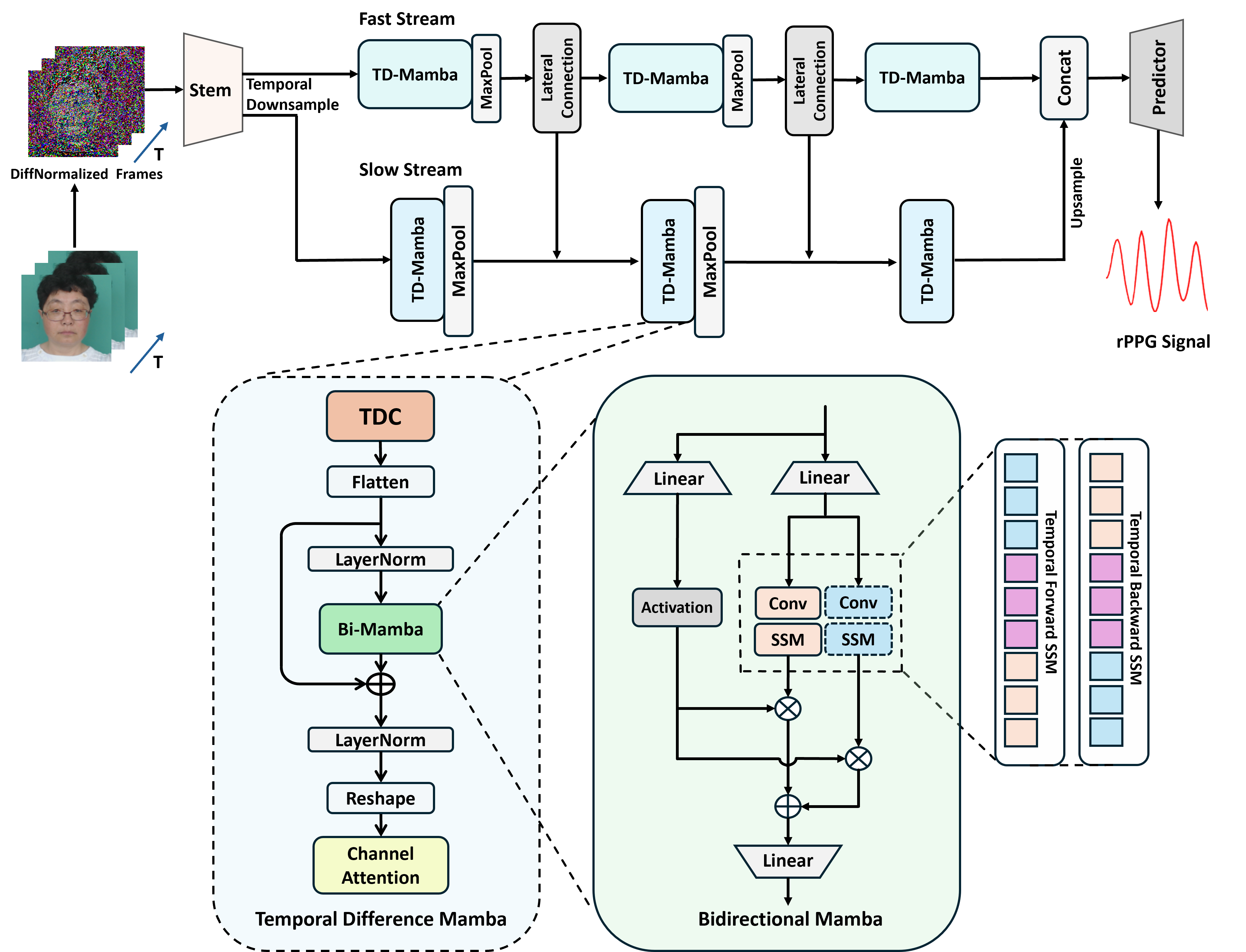}
\vspace{-1.8em}
\caption{\small{Framework of the PhysMamba. It has a shallow stem and a temporal downsample operation ahead. Then for both slow and fast streams, it includes temporal difference Mamba blocks, lateral connections and a rPPG predictor head. Temporal Difference Mamba (TD-Mamba) consists of a Temporal Difference Convolution (TDC), a Temporal Bidirectional Mamba (Bi-Mamba) with forward and backward SSM, and a channel attention (CA) module.}}
\label{fig:network}
\vspace{-1.8em}
\end{figure}

\vspace{-1.5em}
\subsection{Network Architecture}
\vspace{-0.5em}
As shown in Fig.~\ref{fig:network}, PhysMamba mainly consists of a shallow stem, three Temporal Difference Mamba (TD-Mamba) blocks and a rPPG predictor head. Firstly, we apply DiffNormalized\cite{Deepphys1} method to the cropped facial frames for extracting inter-frame differences, which are proved to help for robust rPPG recovery under motion and mitigate the impact of background pixels. We use the shallow stem $E_{stem}$ to extract coarse local spatio-temporal features. Specifically, the stem is formed by three convolutional blocks with kernel size (1x5x5), (3x3x3) and (3x3x3), respectively. Each conlolution block is cascaded with a batch normalization (BN) and ReLU, and the first and last blocks are followed by a pooling layer for halving the spatial dimension. Therefore, given an input RGB facial video, the DiffNormalized frames can be represented as $X \in \mathbb{R}^{3 \times T \times H \times W} $. Then, the the stem $E_{stem}$ generates shallow feature maps $X_{stem} \in \mathbb{R}^{C \times T \times H' \times W' } $, where $H' = \frac{H}{4}$ and $W' = \frac{W}{4}$. Subsequently, we utilize two convolution blocks to perform temporal downsampling on $X_{stem}$ for obtaining slow temporal feature maps $X_{slow} \in \mathbb{R}^{C \times T' \times H' \times W' }$ and fast temporal features $X_{fast} \in \mathbb{R}^{\frac{C}{2} \times 2T' \times H' \times W' }$, where the $T' = \frac{T}{4}$ and the channels of fast temporal features are compressed to $\frac{C}{2}$. Then the Slow and the Fast features will be fed into three Temporal Difference Mamba blocks respectively to perform long-term spatial-temporal modeling. Simultaneously, we add a (1x2x2) Maxpool following each of the first two blocks and use the lateral connections to fuse the Fast stream features into the Slow stream as well. We utilize a temporal convolution with kernel size=3x1x1, stride=2x1x1 and paddings=1x0x0 as the lateral connection. Finally, the last temporally upsampled Slow stream features $X_{slow} \in \mathbb{R}^{C \times 2T' \times \frac{H'}{4} \times \frac{W'}{4}} $ and Fast stream features $X_{fast} \in \mathbb{R}^{\frac{C}{2} \times 2T' \times \frac{H'}{4} \times \frac{W'}{4} }$ are concatenated and forwarded to the rPPG predictor, where temporal upsampling, spatially averaging and 1D rPPG signal $Y \in \mathbb{R}^T $ projection are applied to the final features.

\vspace{-1.4em}
\subsection{Temporal Difference Mamba}
\vspace{-0.5em}
Temporal Difference Convolution (TDC)\cite{Physformer} has been demonstrated to efficiently describe fine-grained local spatio-temporal dynamics which are crucial for tracking subtle color changes. We utilize TDC for enhancing temporally normalized frame difference features representation. TDC can be formulated as: 
\vspace{-0.6em}
\begin{equation} \small
\setlength{\belowdisplayskip}{-0.2em}
\text{TDC}(x) = \underbrace{\sum_{p_n \in \mathcal{R}} w(p_n) \cdot x(p_0 + p_n)}_{\text{vanilla 3D convolution}} + \theta \cdot \underbrace{\left(-x(p_0) \cdot \sum_{p_n \in \mathcal{R}'} w(p_n)\right)}_{\text{temporal difference term}},
\end{equation}
where $w$ are learnable weight parameters, $p_0 = (0,0,0)$ indicates the current patio-temporal location, $\mathcal {R}$ represents the sampled local $3\times 3 \times 3$ neighborhood and $\mathcal {R'}$ indicates the local spatial regions in the adjacent time steps. The hyperparameter $\theta \in [0, 1]$ adjust the contribution of temporal difference. We build a TDC layer cascaded with a batch normalization (BN) and ReLU to extract fine-grained local temporal difference feature maps based on subtle temporal skin color changes. 
Then the spatio-temporal feature maps with a shape of $(B, C, T, H, W)$ are flattened and transposed to 1D long sequence with the size of $(B, L, C)$, where $L = T\times H \times W$. The flattened sequence will be first fed into a LayerNorm(LN), then they can be forwarded to a layer of Temporal Bidirectional Mamba (Bi-Mamba) which can capture inter-frame long-range spatio-temporal dependencies. The procedure in the Mamba layer can be formulated as:
\vspace{-0.6em}
\begin{align} \small
\setlength{\belowdisplayskip}{-0.2em}
h_{k+1} = \text{Bi-Mamba} \left( \text{LN} \left( h_{k} \right) \right) + h_{k},
\end{align}
where the $h_k \in \mathbb{R}^{B \times L \times C}$ denotes the flattened sequence and Bi-Mamba is the Mamba layer with temporal forward and backward SSM. 

Within the Bi-Mamba layer, $h_k$ will be first linearly projected to the hidden states $x$ and $z$ with an expansion factor $E$. Afterward, the $x$ can be flipped along the temporal direction to obtain temporal backward sequence, and then both forward and backward direction sequences can be parallel processed. For each direction, Mamba utilize the 1-D convolution cascaded with the SiLU to the $x$, then it is linearly projected to the $B$, $C$ and $\Delta$. Then the $\Delta$ is used to obtain $\bar B$ and transform parameter $A$ to $\bar A$, and Mamba can performs the core SSM operation with $\bar A$, $\bar B$, $C$ and $x$. At last, the output from both temporal forward and backward direction will be gated by $z$ which is also activated by Silu, and then they are added for the final out put sequence $h_{k+1}$. Subsequently, we use another LayerNorm to normalize the Bi-Mamba output $h_{k+1}$ and transform its shape back to $(B, C, T, H, W)$. Finally, we utilize a Channel Attention (CA) at the end of block to enhance the channel representation. Please find the algorithm pseudocode of the TD-Mamba block in Supplementary Materials. 




\vspace{-1.1em}
\subsection{Loss Function}
\vspace{-0.2em}
We utilize the negative Pearson (NegPearson) loss\cite{PhysNet} as our loss function. The NegPearson loss ensures that the predicted rPPG signals align with the temporal patterns of the ground truth signals, significantly enhancing rPPG signal recovery, where accurate timing and pattern recognition are essential for reliable heart rate monitoring. The NegPearson loss can be defined as:
\vspace{-0.4em}
\begin{equation} \small
\text{Loss} = 1 - \frac{T \sum_{t=1}^{T} x_t y_t - \sum_{t=1}^{T} x_t \sum_{t=1}^{T} y_t}{\sqrt{T \sum_{t=1}^{T} x_t^2 - \left(\sum_{t=1}^{T} x_t \right)^2} \sqrt{T \sum_{t=1}^{T} y_t^2 - \left(\sum_{t=1}^{T} y_t \right)^2}},
\end{equation}
where $T$ is the length of the signals, $x$ represents the predicted rPPG signals, and $y$ denotes the ground truth rPPG signals. 

\vspace{-1.4em}
\section{Experiments}
\vspace{-0.6em}
\subsection{Dataset and Metrics}
\vspace{-0.3em}
We use three benchmark datasets: \textbf{PURE}\cite{PURE}, \textbf{UBFC-rPPG}\cite{UBFC} and \textbf{MMPD}\cite{MMPD} for evaluation. The PURE dataset comprises 60 one-minute videos from 10 subjects (8 males, 2 females) performing six different activities, with a frame rate of 30Hz and a resolution of 640 × 480. UBFC-rPPG includes 42 facial videos from participants who were asked to engage in a time-sensitive mathematical game. The videos are captured at 30fps with a resolution of 640x480. MMPD consists of 660 one-minute videos with a resolution of 320x240 and a frame rate of 30Hz from 33 subjects with diverse skin types and activities under four lighting conditions. MMPD provides a compressed version dataset named mini-MMPD. We use the mini-MMPD version in our experiments. For evaluation metrics, Mean Absolute Error (MAE), Root Mean Squared Error (RMSE), Mean Absolute Percentage Error (MAPE) and Pearson's correlation coefficient ($\rho$) are used for HR estimation evaluation. HR is measured in beats per minute (bpm).

\vspace{-1.3em}
\subsection{Implementation Details}
\vspace{-0.5em}
We conduct experiments on Pytorch and mainly based on the open-source toolkit rPPG-Toolbox\cite{rppg-toolbox}. For data pre-processing, we crop the face region in the first frame for each video clip and fix the region box in the following frames. Subsequently, we randomly sample a video chunk of 128 frames and resize them into 128 $\times$ 128 pixels. In terms of DiffNormalized\cite{Deepphys1}, the difference between two frames is first calculated by $(X_{t+1} - X_{t}/X_{t} + X_{t+1})$, and then they are normalizes by their standard deviation. The channels of slow and fast streams are 64 and 32, respectively. We use the default hyperparameters settings N = 16 and E = 2 of Mamba and choose $\theta = 0.5$ for TDC. The PhysMamba is trained with Adam optimizer with learning rate of 3e-3 and weight decay of 5e-4. We train our model for 20 epochs on a NVIDIA RTX 4090 GPU with batch size of 4. 


\vspace{-1.3em}
\subsection{Intra-dataset Evaluation}
\vspace{-0.3em}
UBFC-rPPG and PURE datasets are used for intro-dataset test on HR estimation task. We followed\cite{PURE64} to use 36 videos of the PURE dataset for training and 24 videos for testing. For the evaluation on UBFC-rPPG dataset, we followed\cite{PulseGAN} to use the initial 30 samples for training and the remaining 12 samples for testing. As shown in Table~\ref{tab:intro}, PhysMamba achieves the lowest MAE(0.25 bpm), RMSE (0.4bpm) on the PURE dataset and exhibits comparable performance with state-of-the-art method PhysFormer\cite{Physformer} on the UBFC-rPPG dataset, indicating the effectiveness of the design of the SlowFast based TD-Mamba framework.

\begin{table}[t]
\centering
\vspace{-1.9em}
\caption{\small{Intra-dataset testing results on \textbf{PURE} and \textbf{UBFC-rPPG}.}}
\resizebox{0.68\textwidth}{!}{\begin{tabular}{lcccccc}
\toprule
\multirow{3}{*}{\textbf{Method}} & \multicolumn{3}{c}{\textbf{PURE}} & \multicolumn{3}{c}{\textbf{UBFC-rPPG}} \\
\cmidrule(lr){2-4} \cmidrule(lr){5-7}
& MAE $\downarrow$ & RMSE $\downarrow$ & $\rho$ $\uparrow$ & MAE $\downarrow$ & RMSE $\downarrow$ & $\rho$ $\uparrow$ \\
\midrule
TS-CAN\cite{TS-CAN} & 2.48 & 9.01 & 0.99 & 1.70 & 2.72 & 0.99 \\
PhysNet\cite{PhysNet} & 2.10 & 2.60 & 0.99 & 2.95 & 3.67 & 0.97 \\
DeepPhys\cite{Deepphys1}  & \underline{0.83} & \underline{1.54} & 0.99 & 6.27 & 10.82 & 0.65 \\
EfficientPhys\cite{Efficientphys1}  & - & - & - & 1.14 & 1.81 & 0.99 \\
PhysFormer\cite{Physformer} & 1.10 & 1.75 & 0.99 & \textbf{0.50} & \textbf{0.71} & \textbf{0.99} \\
\midrule
\textbf{PhysMamba (Ours)} & \textbf{0.25} & \textbf{0.4} & \textbf{0.99} & \underline{0.54} & \underline{0.76} & \textbf{0.99} \\
\bottomrule
\end{tabular}}
\label{tab:intro}
\vspace{-1.8em}
\end{table}

\begin{table}[t]
\centering
\caption{\small{Cross-dataset results training on \textbf{UBFC-rPPG}.}}
\resizebox{0.68\textwidth}{!}{\begin{tabular}{lcccccc}
\toprule
\multirow{3}{*}{\textbf{Method}} & \multicolumn{3}{c}{\textbf{PURE}} & \multicolumn{3}{c}{\textbf{MMPD}} \\
\cmidrule(lr){2-4} \cmidrule(lr){5-7}
& MAE $\downarrow$ & RMSE $\downarrow$ & $\rho$ $\uparrow$ & MAE $\downarrow$ & RMSE $\downarrow$ & $\rho$ $\uparrow$ \\
\midrule
TS-CAN\cite{TS-CAN} & 3.69 & 13.8 & 0.82 & 14.01 & 21.04 & 0.24 \\
PhysNet\cite{PhysNet} & 8.06 & 19.71 & 0.61 & \textbf{9.47} & \textbf{16.01} & \textbf{0.31} \\
DeepPhys\cite{Deepphys1}  & 5.54 & 18.51 & 0.66 & 17.50 & 25.00 & 0.05 \\
EfficientPhys\cite{Efficientphys1}  & 5.47 & 17.04 & 0.71 & 13.78 & 22.25 & 0.09 \\
PhysFormer\cite{Physformer} & 12.92 & 24.36 & 0.47 & 12.1 & 17.79 & 0.17 \\
SpikingPhys\cite{Spiking-physformer} & \underline{2.70} & - & \underline{0.91} & 13.36 & - & 0.20 \\
\midrule
\textbf{PhysMamba (Ours)} & \textbf{1.20} &\textbf {5.99} & \textbf{0.97} & \underline{11.96} & \underline{17.69} & \underline{0.29} \\
\bottomrule
\end{tabular}}
\label{tab:traning on UBFC}
\vspace{-2.3em}
\end{table}


\vspace{-1.3em}
\subsection{Cross-dataset Evaluation}
\vspace{-0.3em}
We followed protocols in rPPG-Toolbox\cite{rppg-toolbox} for the cross-dataset evaluation. The training datasets are divided into 8:2 for training and validation. We conducted the experiments by training on the PURE and UBFC-rPPG datasets, and evaluated the HR estimation on the PURE, UBFC-rPPG, and MMPD datasets. The results are shown in Table~\ref{tab:traning on UBFC} and Table~\ref{tab:training on PURE}. The Proposed PhysMamba achieves the lowest MAE (1.20bpm), RMSE (5.99bpm) and highest $\rho$ (0.97) on the PURE dataset when training on the UBFC-rPPG dataset, and it shows state-of-the-art performance on the both UBFC-rPPG and MMPD datasets when training on the PURE dataset. It can be seen that testing results on the MMPD dataset are much lower than those on the other two datasets, since the environment and subjects in the MMPD datasets are much more diverse and complex. 

\begin{figure}[t]
\centering
\includegraphics[width=0.8\textwidth]{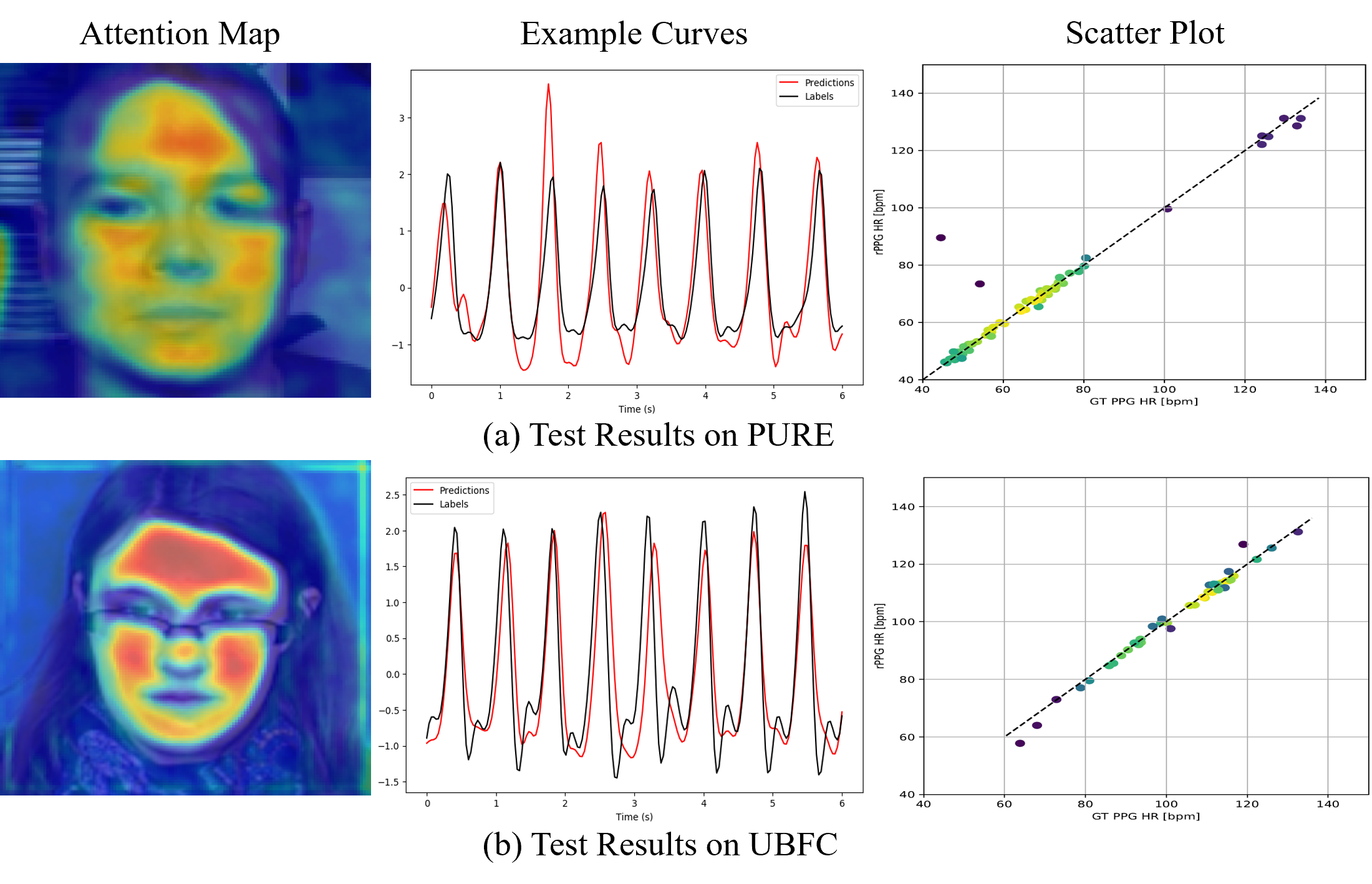}
\vspace{-1.2em}
\caption{\small{Attention map, example curves of predicted rPPG signals with ground truth and Scatter Plot of cross-dataset results testing on (a) PURE and (b) UBFC-rPPG.}}
\label{fig:visuals}
\vspace{-1.6em}
\end{figure}

We also provide visualizations with testing on PURE and UBFC-rPPG to demonstrate the superior performance of our model. As shown in Fig.~\ref{fig:visuals}, the attention maps can demonstrate that our model can effectively focus on facial regions especially the forehead and cheeks with rich hemoglobin. In addition, We also present example rPPG signal curves and Scatter plots of HR, which can indicate the strong correlation between the ground truth and our predictions.

\begin{table}[t]
\vspace{-1.8em}
\centering
\caption{\small{Cross-dataset results training on \textbf{PURE}.}}
\resizebox{0.68\textwidth}{!}{\begin{tabular}{lcccccc}
\toprule
\multirow{3}{*}{\textbf{Method}} & \multicolumn{3}{c}{\textbf{UBFC-rPPG}} & \multicolumn{3}{c}{\textbf{MMPD}} \\
\cmidrule(lr){2-4} \cmidrule(lr){5-7}
& MAE $\downarrow$ & RMSE $\downarrow$ & $\rho$ $\uparrow$ & MAE $\downarrow$ & RMSE $\downarrow$ & $\rho$ $\uparrow$ \\
\midrule
TS-CAN\cite{TS-CAN} & 1.30 & 2.87 & 0.99 & 13.94 & 21.61 & 0.20 \\
PhysNet\cite{PhysNet} & \underline{0.98} & \underline{2.48} & \underline{0.99} & 13.93 & 20.29 & 0.17 \\
DeepPhys\cite{Deepphys1}  & 1.21 & 2.90 & 0.99 & 16.92 & 24.61 & 0.05 \\
EfficientPhys\cite{Efficientphys1}  & 2.07 & 6.32 & 0.94 & 14.03 & 21.62 & 0.17 \\
PhysFormer\cite{Physformer} & 1.44 & 3.77 & 0.98 & 14.57 & 20.71 & 0.15 \\
SpikingPhys\cite{Spiking-physformer} & 5.25 & - & 0.83 & \underline{12.76} & - & \underline{0.23} \\
\midrule
\textbf{PhysMamba (Ours)} & \textbf{0.97} & \textbf{1.93} & \textbf{0.99} & \textbf{10.31} & \textbf{16.02} & \textbf{0.34} \\
\bottomrule
\end{tabular}}
\label{tab:training on PURE}
\vspace{-1.1em}
\end{table}

\vspace{-1.3em}
\subsection{Ablation Study}
\vspace{-0.5em}
In addition, we provide ablation studies for HR estimation on the UBFC-rPPG dataset. We investigated the impact of key modules in the TD-Mamba block and SlowFast architecture. It is shown in Table~\ref{tab:td_mamba} that the exclusion of both Temporal Difference Convolution (TDC) and Mamba results in a performance decline. Without temporal difference convolution, the performance sharply drop to MAE(0.68bpm), RMSE (0.97bpm) and MAPE (0.69), indicating that fine-grained temporal difference features are significant for the rPPG signals modeling even though the Mamaba capture the long-range spatio-temporal dependencies. Additionally, as can be seen in Table~\ref{tab:slowfast}, the ablation of the Slow stream also pose a dramatic decline to HR estimation performance. The fusion spatio-temporal features in Slow and Fast streams significantly improve performance.


\begin{table}[t]
\vspace{-1.9em}
\centering
\begin{minipage}[t]{0.45\textwidth}
\centering
\caption{\small{Ablation of TD-Mamba.}}
\resizebox{0.78\textwidth}{!}{\begin{tabular}{lccc}
\toprule
\multirow{2}{*}{\textbf{Method}} & \multicolumn{3}{c}{\textbf{Test on UBFC-rPPG}} \\
\cmidrule(lr){2-4}
& MAE $\downarrow$ & RMSE $\downarrow$ & MAPE $\downarrow$ \\
\midrule
w/o TDC & 0.68 & 0.97 & 0.69 \\
w/o Mamba & 0.63 & 0.85 & 0.65 \\
w/o CA & 0.62 & 0.84 & 0.63 \\
w/o Bi-SSM & 0.59 & 0.84 & 0.60 \\
\midrule
\textbf{Ours} & \textbf{0.54} & \textbf{0.76} & \textbf{0.56} \\
\bottomrule
\end{tabular}}
\label{tab:td_mamba}
\end{minipage}
\hspace{0.08\textwidth}
\begin{minipage}[t]{0.45\textwidth}
\centering
\caption{\small{Ablation of SlowFast fusion.}}
\resizebox{0.88\textwidth}{!}{\begin{tabular}{lccc}
\toprule
\multirow{2}{*}{\textbf{Method}} & \multicolumn{3}{c}{\textbf{Test on UBFC-rPPG}} \\
\cmidrule(lr){2-4}
& MAE $\downarrow$ & RMSE $\downarrow$ & MAPE $\downarrow$ \\
\midrule
Slow-only & 0.63 & 0.85 & 0.65 \\
Fast-only & 0.81 & 1.05 & 0.82 \\
w/o Lateral Connect & 0.59 & 0.84 & 0.60 \\
\midrule
\textbf{Ours} & \textbf{0.54} & \textbf{0.76} & \textbf{0.56} \\
\bottomrule
\end{tabular}}
\label{tab:slowfast}
\end{minipage}
\vspace{-2.3em}
\end{table}

\begin{wraptable}{r}{6.2cm}
\centering
\caption{\small{Comparison on Param. and MACs}}
\vspace{-0.3cm}
\resizebox{0.45\textwidth}{!}{\begin{tabular}{lcc}
\toprule
\textbf{Method} & \textbf{Param. (M)} & \textbf{MACs (G)} \\
\midrule
TS-CAN\cite{TS-CAN} & 7.5 & 96 \\
PhysNet\cite{PhysNet} & 0.77 & 56.1 \\
DeepPhys\cite{Deepphys1} & 7.5 & 96 \\
EfficientPhys\cite{Efficientphys1} & 7.4 & 45.6 \\
PhysFormer\cite{Physformer} & 7.38 & 40.5 \\
\midrule
\textbf{PhysMamba(Ours)}  & \textbf{0.56} & 47.3 \\
\bottomrule
\end{tabular}}
\label{tab:comparison}
\end{wraptable}

\vspace{-1.0em}
\subsection{Parameters and Computational Efficiency}
\vspace{-0.4em}
In Table 6, we compare the number of parameters and multiply-accumulate operations (MACs) with other different models. PhysMamba effectively reduces the number of parameters to 0.56M, while maintaining relatively low computational complexity with 47.3G MACs with the input size of $128 \times 128 \times 128  (T \times H \times W)$, exhibiting the potential for deployment in resource-constrained mobile devices.

\vspace{-1.3em}
\section{Conclusion}
\vspace{-0.7em}
In this paper, we propose a Mamba-based model PhysMamba for remote physiological measurement. Specifically, the Temporal Difference Mamba (TD-Mamba) block and dual-stream SlowFast architecture are introduced to enhance the extraction of spatio-temporal features for efficient rPPG signals modeling. Experiments conducted on three benchmark datasets demonstrate PhysMamba's superior performance compared with existing deep learning methods. 
\vspace{-1.5em}
\subsubsection*{Acknowledgments.} This work was supported by National Natural Science Foundation of China under Grant 62306061.


\bibliographystyle{abbrv}
\vspace{-1.0em}
\bibliography{references}

\section{Algorithm of TD-Mamba Block}
\begin{algorithm}[h]
\caption{Temporal Difference Mamba Block}
\begin{algorithmic}[1]

\REQUIRE $f: (B, C, T, H, W)$
\ENSURE $f': (B, C, T, H, W)$
\STATE \quad $f: (B, C, T, H, W)\leftarrow \text{TDC}(f) $  
\STATE \quad $f: (B, C, T, H, W)\leftarrow  \text{ReLU} (\text {BN} (f))$ 

\STATE \quad $h_{k}: (B, L, C)\leftarrow  \text {Flatten} (f)$ 

\STATE \quad $x, z: (B, L, E)\leftarrow \text{Linear}(\text{LN}(h_{k}))$ 
\FOR{direction in \{forward, backward\}}
    \IF{direction = backward}
        \STATE \quad $x \leftarrow$ Flip(x, dims = 1) 
    \ENDIF
    \STATE \quad $x: (B, L, E) \leftarrow \text{SiLU}(\text{Conv1d}(x))$ 
    \STATE \quad $A: (C, N) \leftarrow \text{Parameter}$
    \STATE \quad $B: (B, L, N)\leftarrow \text{Linear}(x)$
    \STATE \quad $C: (B, L, N)\leftarrow \text{Linear}(x)$
    \STATE \quad $\Delta : (B, L, C)\leftarrow \text{SoftPlus}((\text{Parameter})+s_\Delta(x)$ 
    \STATE \quad $\bar A : (B, L, C, N)\leftarrow \text{Parameter}_A \otimes \Delta$
    \STATE \quad $\bar B : (B, L, C, N)\leftarrow B \otimes \Delta$
    \STATE \quad $y : (B, L, E)\leftarrow \text{SSM}(\bar A, \bar B, C )(x)$
\ENDFOR
\STATE \quad $y_{\text{forward}}: (B, L, E) \leftarrow y_{\text{forward}} \odot \text{SiLU}(z)$
\STATE \quad $y_{\text{backward}}: (B, L, E) \leftarrow y_{\text{backward}} \odot \text{SiLU}(z)$


\STATE \quad $h_{k+1}: (B, L, C)\leftarrow \text{Linear}(y_{forward} + y_{backward}) + h_{k}$

\STATE \quad $g: (B, C, T, H, W)\leftarrow \text{Reshape}(\text {LN}(h_{k+1}))$ 

\STATE \quad $f‘: (B, C, T, H, W)\leftarrow \text{CA}(g)$

\RETURN $f'$

\end{algorithmic}
\end{algorithm}

\end{document}